\documentclass[11pt,a4paper]{article}
\usepackage{naaclhlt2018}
\usepackage{times}
\usepackage{latexsym}

\usepackage{url}

\aclfinalcopy % Uncomment this line for the final submission
\usepackage{booktabs}
\usepackage{lipsum}
\usepackage{bbm}

\usepackage{chngpage}
\usepackage{stmaryrd}
\usepackage{amssymb}
\usepackage{amsmath}
\usepackage{amsthm}
\usepackage{graphicx}
\usepackage{lscape}
\usepackage{subfigure}
\usepackage{color}
\definecolor{myblue}{rgb}{0,0.1,0.6}
\definecolor{mygreen}{rgb}{0,0.3,0.1}
\newcommand{\ignore}[1]{}

\title{Monte Carlo Syntax Marginals for Exploring and Using \\Dependency Parses}

\author {Katherine A. Keith, Su Lin Blodgett, \textnormal{and} Brendan O'Connor \\
College of Information and Computer Sciences \\
 University of Massachusetts Amherst \\
 {\tt \{kkeith,blodgett,brenocon\}@cs.umass.edu} 
 }

\date{}

\begin{document}
\maketitle
\begin{abstract}
Dependency parsing research, which has made significant gains in recent years, typically focuses on improving the accuracy of single-tree predictions. However, ambiguity is inherent to natural language syntax, and communicating such ambiguity is important for error analysis and better-informed downstream applications. In this work, we propose a \emph{transition sampling}
 algorithm to sample from the full joint distribution of parse trees
defined by a transition-based parsing model,
and demonstrate the use of the samples in probabilistic dependency analysis.
First, we define the new task of \emph{dependency path prediction}, inferring syntactic substructures 
over part of a sentence, and provide the first analysis of performance on this task.
Second, we demonstrate the usefulness of our \emph{Monte Carlo syntax marginal} method for parser error analysis and calibration.
Finally, we use this method to propagate parse uncertainty to two downstream information extraction applications: identifying persons killed by police and semantic role assignment.\footnote{Supporting code available at \url{https://github.com/slanglab/transition_sampler}.}
\medbreak 
[This paper appears in \emph{Proceedings of NAACL 2018}]
\end{abstract}

\section{Introduction} \label{s:intro}

\newcommand{\telsfigure}{
\begin{figure}[t]
\centering
\includegraphics[width=.8\linewidth]{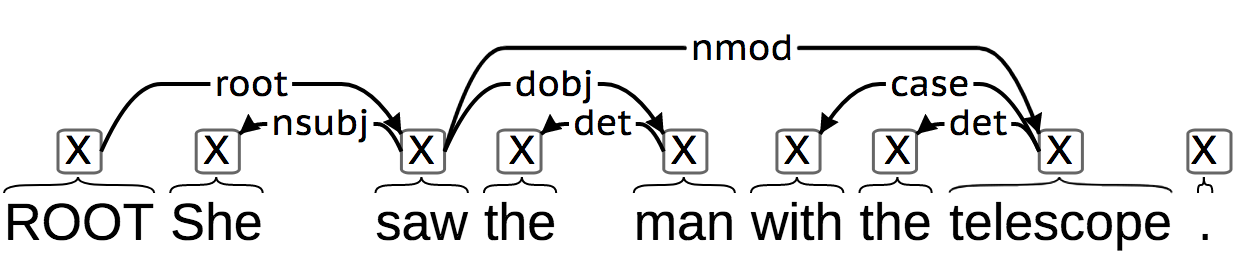}
\includegraphics[width=1\linewidth]{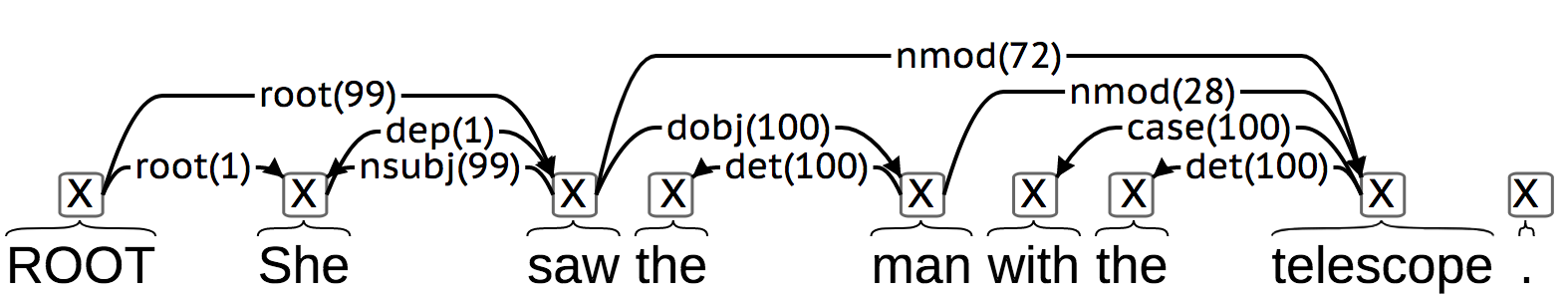}
\caption{Example of a sentence with inherent ambiguity. \textbf{Top:} output from a greedy parser. \textbf{Bottom:} edge marginal probabilities from 100 samples in parentheses.}
\label{tels}
\end{figure}}

\noindent
Dependency parsers typically predict a single tree for a sentence to be used in downstream applications,
and most work on dependency parsers seeks to improve accuracy of such single-tree predictions.
Despite tremendous gains in the last few decades of parsing research,
accuracy is far from perfect---but perfect accuracy may be impossible
since syntax models by themselves do not
incorporate the discourse, pragmatic, or world knowledge
necessary to resolve many ambiguities.

In fact, although relatively unexamined, substantial ambiguity
already exists within commonly used
discriminative probabilistic parsing models,
 which define a parse forest---a 
 probability distribution $p(y \mid x)$ over possible dependency trees $y \in \mathcal{Y}(x)$
 for an input sentence $x$.
 
 \telsfigure
  
For example, the top of Figure~\ref{tels} shows the predicted parse $y^{(greedy)}$ from
such a parser \citep{Chen2014NN},
% transition-based neural dependency parser,\footnote{From the version in CoreNLP 3.8.0}
which resolves a prepositional (PP) attachment ambiguity in one manner;
this prediction was selected by a standard greedy transition-based algorithm (\S\ref{s:parsemodel}).
However, the bottom of Figure~\ref{tels} shows \emph{marginal} probabilities of individual (relation, governor, child)
edges under this same model.
These denote our estimated probabilities, across all possible parse structures, that a pair of words
are connected with a particular relation (\S\ref{s:squery}).
For example, the two different PP attachment readings
both exist within this parse forest with marginal probabilities
\begin{align}
p( \text{ nmod}(\text{saw}_2,\text{telescope}_7) \mid x)&=0.72 \\
p( \text{ nmod}(\text{man}_4,\text{telescope}_7) \mid x)&=0.28,
\end{align}
where (1) implies she used a telescope to see the man, and (2) implies she saw a man who had a telescope. 

These types of irreducible syntactic ambiguities exist and 
should be taken into consideration when analyzing syntactic information;
for instance, one could transmit multiple samples \cite{Finkel2006Pipeline}
or
confidence scores \cite{Bunescu2008Pipeline}
over ambiguous readings to downstream analysis components.

In this work, we introduce
a simple \emph{transition sampling} algorithm for transition-based dependency parsing
(\S\ref{s:ancestral}), which, by yielding exact samples from the full joint distribution over trees,
makes it possible to infer probabilities of long-distance or other arbitrary structures over the 
parse distribution (\S\ref{s:squery}). We implement transition sampling---a very simple
change to pre-existing parsing software---and use
it to demonstrate several applications of probabilistic dependency analysis:
\begin{itemize}
\itemsep0em
\item Motivated by how dependency parses are typically used in feature-based machine learning, 
we introduce a new parsing-related task---\emph{dependency path prediction}. This task involves inference over variable length \emph{dependency paths}, syntactic substructures over only parts of a sentence.
\item To accomplish this task, we define a \emph{Monte Carlo syntax marginal} inference method
which exploits information across samples of the entire parse forest.
It achieves higher accuracy predictions
than a traditional greedy parsing algorithm,
and allows tradeoffs between precision and recall (\S\ref{s:pr}).
\item We provide a quantitative measure of the model's inherent uncertainty in the parse, \emph{whole-tree entropy}, and show how it can be used for error analysis
(\S\ref{s:eda}).
\item  We demonstrate the method's (surprisingly) reasonable calibration (\S\ref{s:calib}).
\item Finally, we demonstrate the utility of our method to propagate uncertainty to downstream applications. Our method improves performance for
giving probabilistic semantics to a rule-based event extractor to identify civilians killed by police (\S\ref{s:rules}),
as well as semantic role assignment (\S\ref{ss:srl}).
\end{itemize}

\section{Monte Carlo dependency analysis} \label{s:model}

%\bocomment{this section describes the ancestral sampling method}
%
%\subsection*{random notes, move things around}
%
%How well does greedy decoding find the mode?
%It disagrees with MC-MAP a good bit.  MC-MAP shows greedy decoding could be improved?
%
%idea: take unique structures from MC samples and score them with their calculated joint probs.  in expectation this is the same as their frequencies in the MC sample.  but perhaps has less variance in estimating prob weights?  there must be a name for this already.
%
\subsection{Overview of transition-based dependency parsing} \label{s:parsemodel}

We examine the \emph{basic} form of the
Universal Dependencies formalism \cite{Nivre2016UD},
where, for a sentence $x$ of length $N$,
a possible dependency parse $y$ is a set of
(relation, governorToken, childToken) edges,
with a tree constraint that every token in the parse has exactly one governor---that is,
for every token $w \in \{1..N\}$, there is exactly one triple $(r,g,w) \in y$
where it participates as a child. The governor is either one of the observed tokens, or a special ROOT vertex.
%$g \in \{ROOT\} \cup \{1..N\}$.

There exist a wide variety of approaches to machine learned, discriminative dependency parsing,
which often define a probability distribution $p( y \mid x)$ over 
a domain of formally legal dependency parse trees $y \in \mathcal{Y}(x)$.
We focus on \emph{transition-based}
dependency parsers \cite{Nivre2003Efficient,Kubler2009Deps},
which (typically) use a stack-based automaton
to process a sentence, incrementally building a set of edges.
Transition-based parsers are very fast, have runtimes linear in sentence length,
feature high performance (either state-of-the-art, or nearly so),
and are easier to implement than other modeling paradigms
(\S\ref{relatedwork}).

A probabilistic transition-based parser assumes
the following stochastic process
to generate a parse tree:

\fbox{\begin{minipage}[t]{2.5in} 
\begin{itemize}
\itemsep0em
\item Initialize state $S_0$ %=(\text{Stack}=[], \text{Buf}=x, \text{Arcs}=\{\})$
\item For $n=1, 2, \ldots$:
\begin{enumerate}
%\item[(A)] Choose $a_n$
\item[(A)] $a_n \sim p( a_n \mid S_{n-1} )  $
\item[(B)] $S_n: = \text{Update}(S_{n-1}, a_n)$
\item[(C)] Break if InEndState($S_n$)
\end{enumerate}
\end{itemize}

\end{minipage}}

\noindent
Most state transition systems \citep{Bohnet2016Transition}
use shift and reduce actions
to sweep through tokens from left to right, pushing and popping them from a stack to create the edges
that populate a new parse tree $y$.
The action decision probability, $p(a_{next} \mid S_{current})$, is a softmax distribution over possible
next actions.  It can be parameterized
 by any probabilistic model,
 such as
log-linear features of the sentence and current state \cite{Zhang2011Transition},
multilayer perceptrons \cite{Chen2014NN}, 
or
recurrent neural networks \cite{Dyer2015Deps,Kiperwasser2016Parsing}.
%The action model is typically trained to maximize the model's
%log-likelihood of oracle action sequences derived from a gold-standard treebank.

%(other similar models... Andors .. etc)
%To train the model, a gold standard dependency treebank is converted into oracle action sequences that could have generated the observed gold-standard parsers; the parameters $\theta$ (e.g.\ MLP weights)
%are trained to maximize the log-likelihood of oracle actions.

To predict a single parse tree on new data, a common inference method is \emph{greedy decoding},
which runs a close variant of the above transition model
as a deterministic automaton, 
replacing stochastic step (A) with
a best-action decision,
$a_n := \arg\max_{a_n} p( a_n \mid S_{n-1} )$.\footnote{Since greedy parsing does not require probabilistic semantics for the action model---the softmax normalizer does not need to be evaluated---non-probabilistic 
training, such as with hinge loss (SVMs), is a common alternative,
including in some of the cited work.}
An inferred action sequence $a_{1:n}$ determines the resulting parse tree (edge set) $y$;
the relationship can be denoted as $y(a_{1:n})$.

%Work has also sought to improve the selection of the parse with beam search, dynamic programming; and analogous work
%
%This is used by \citet{Chen2014NN} (..and others?..),
%and also commonly with a non-probabilistic scoring function for action choice,
%e.g. SVMs (Nivre) or CRF-style,
%globally normalized scoring functions (Andors) which only have probabilistic semantics 
%at the level of the entire structure.  But here we focus on models with locally normalized probabilistic decisions.
%
%The probability of an entire action sequence 
%$(a_1..a_n)$
%(which determines one particular structure)
%is therefore
%\[ p(a_1 .. a_n \mid x) = \prod_{i=1..n} p(a_i \mid a_{<i}, x) \]
%Much research has noted that it may be desirable to instead try to choose a parse structure
%(or more specifically, an action sequence) that maximizes this joint probability, instead of greedily making local choices at each timestep.  Heuristic search algorithms such as beam search (..cite..),
%or many variants (choi and mccallum, ?, balleseros, dyer...) are typically used.
%(also constit parsing cites: zhang/clarke, roark/collins, dyer RNNG, ?+klein on choe/charniak version of RNNG)

\subsection{Transition sampling} \label{s:ancestral}

In this work, we propose to analyze the full joint posterior $p(y \mid x)$,
and use \emph{transition sampling}, 
a very simple forward/ancestral sampling algorithm,\footnote{``Ancestral'' refers to a directed Bayes net (e.g.\ \citet{Barber2012BRML})
of action decisions, each conditioned on the full history of previous actions---not ancestors in a parse tree.}
to draw parse tree samples from that distribution.
To parse a sentence, we run the automaton stochastically,
sampling the action probability in step (A).
This yields one action sequence $a_{1:n}$
from the full joint distribution of action sequences,
and therefore a parse $y(a_{1:n})$ from the distribution of parses.
We can obtain as many parse samples
as desired by running the transition sampler $S$ times,
yielding a collection (multiset) of parse structures $\{y^{(s)} \mid s \in \{1..S\}\}$,
where each $y^{(s)} \sim p(y \mid x)$ is a full dependency parse tree.\footnote{\citet{Dyer2016RNNG} use
the same algorithm
to draw samples from a transition-based constituency parsing model,
as an importance sampling proposal to support
parameter learning and single-tree inference.
}
Runtime to draw one parse sample is very similar to the greedy algorithm's runtime.
We denote the set of unique parses
in the sample
$\tilde{\mathcal Y}(x)$.

We implement a transition sampler by modifying
 an implementation of \citeauthor{Chen2014NN}'s
 multilayer perceptron
transition-based parser\footnote{CoreNLP 3.8.0 with its `english\_UD' pretrained model.}
and use it for all subsequent experiments.

\subsection{MC-MAP single parse prediction} \label{s:mcmap}

\noindent
One minor use of transition sampling
is a method for predicting a single parse,
by selecting the most probable (common) parse tree in the sample,
\begin{align}
\hat{y}^{\text{MC-MAP}} &= \arg\max_{y \in \tilde{\mathcal Y}}\   \tilde{p}(y\mid x)
\\
&= \arg\max_{y \in \tilde{\mathcal Y}}\   \frac{c(y)}{S}
\end{align}
where $\tilde{p}(y \mid x)$ denotes the Monte Carlo estimate
of a parse's probability,
which is proportional to how many times it appears in the sample:
$c(y) \equiv \sum_{s}^S 1\{y=y^{(s)}\}$.
Note that $\tilde{p}(y \mid x)$ correctly accounts for the case of
an ambiguous transition system where multiple different action sequences
can yield the same tree---i.e., $y(a_{1:n})$ is not one-to-one---since the transition sampler
can sample the multiple different paths.

%\[y^{MAP} = \arg\max_{y \in \mathcal{Y}(x)} p(y \mid x) \]
This ``MC-MAP'' method is asymptotically
guaranteed to find the model's 
most probable parse ($\arg\max_y p(y \mid x)$)
given enough samples.\footnote{This holds since
the Monte Carlo estimated probability of any tree
converges to its true probability, according to, e.g., Hoeffding's inequality or the central limit theorem.
Thus, with enough samples, the tree with the highest true probability
will have estimated probability higher than any other tree's.}
%so the most probable tree will have a higher estimated
%probability than all other trees.
By contrast, greedy decoding and beam search have no theoretical guarantees.
MC-MAP's disadvantage is that it may require a large number of samples, depending on
the difference between the top parse's probability compared to other parses in the domain. 
%We leave more in-depth investigation for future work.
%\bocomment{cite particlefilters here, or elsewhere}

\newcommand{\prfigure}{
\begin{figure*}[ht]
\scalebox{0.95}{
\begin{minipage}{3.4in}
%\hspace{0.6in}
%\vspace{0.2in}
\includegraphics[width=\textwidth,trim=0 0 0 .8in]{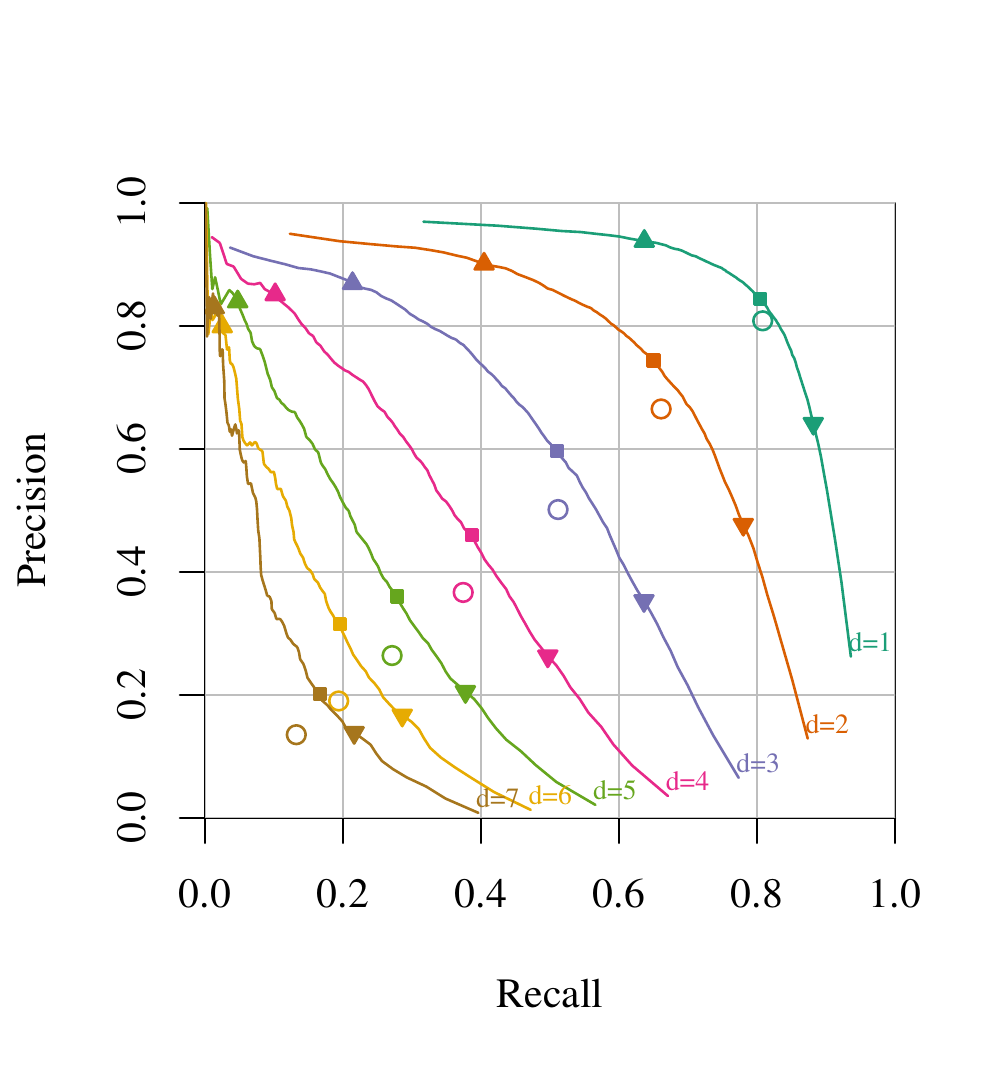}
\end{minipage}
\hspace{-0.05in}
\begin{minipage}{3in}
\vspace{-0.6in}
\centering
\begin{tabular}{@{}ccccc@{}}
\toprule
Path & Greedy & 
\multicolumn{2}{c}{Marginal} & MC-MAP \\
Len. &
\footnotesize{F1} & 
\footnotesize{Max F1} & 
\footnotesize{Thresh.} &
\footnotesize{F1} 
\\
\cmidrule(r){1-1} \cmidrule(lr){2-2} \cmidrule(lr){3-4} \cmidrule(l){5-5} 
\textcolor[HTML]{1B9E77}{1} & \textcolor[HTML]{1B9E77}{$\bigcirc$} 0.808 & \textcolor[HTML]{1B9E77}{$\blacksquare$} 0.824 & \footnotesize{0.45} & 0.807 \\ \textcolor[HTML]{D95F02}{2} & \textcolor[HTML]{D95F02}{$\bigcirc$} 0.663 & \textcolor[HTML]{D95F02}{$\blacksquare$} 0.694 & \footnotesize{0.44} & 0.660 \\ \textcolor[HTML]{7570B3}{3} & \textcolor[HTML]{7570B3}{$\bigcirc$} 0.506 & \textcolor[HTML]{7570B3}{$\blacksquare$} 0.550 & \footnotesize{0.34} & 0.501 \\ \textcolor[HTML]{E7298A}{4} & \textcolor[HTML]{E7298A}{$\bigcirc$} 0.370 & \textcolor[HTML]{E7298A}{$\blacksquare$} 0.420 & \footnotesize{0.28} & 0.363 \\ \textcolor[HTML]{66A61E}{5} & \textcolor[HTML]{66A61E}{$\bigcirc$} 0.268 & \textcolor[HTML]{66A61E}{$\blacksquare$} 0.314 & \footnotesize{0.25} & 0.262 \\ \textcolor[HTML]{E6AB02}{6} & \textcolor[HTML]{E6AB02}{$\bigcirc$} 0.192 & \textcolor[HTML]{E6AB02}{$\blacksquare$} 0.241 & \footnotesize{0.25} & 0.188 \\ \textcolor[HTML]{A6761D}{7} & \textcolor[HTML]{A6761D}{$\bigcirc$} 0.134 & \textcolor[HTML]{A6761D}{$\blacksquare$} 0.182 & \footnotesize{0.17} & 0.131 \\
\bottomrule
\end{tabular}

\vspace{0.15in}
\fbox{
    \newcommand{\mywidth}{\hspace{0.8em}}
%    $\bigcirc$ Greedy \mywidth
    $\blacktriangle$ Conf$\geq$0.9 \mywidth
    $\blacktriangledown$ Conf$\geq$0.1
%    \raisebox{\depth}{\rotatebox[origin=c]{180}{$\triangle$}} Conf$\geq$0.1
    }
\end{minipage}
}
\vspace{-0.2in}
\caption{\label{f:pr} Precision-recall analysis against gold UD 1.3 English dev (\S\ref{ss:dataset}) set for all dependency paths of lengths 
\textbf{\textcolor[HTML]{1B9E77}{1}} to 
\textbf{\textcolor[HTML]{A6761D}{7}}.
\textbf{Left:} Each path length is a different color 
(\textbf{\textcolor[HTML]{1B9E77}{1}} in top-right, 
\textbf{\textcolor[HTML]{A6761D}{7}} in bottom-left),
with greedy performance ($\bigcirc$) as well as the marginal path predictions'
PR curve, with points at confidence thresholds 
0.9
    ($\blacktriangle$),
    0.1
    ($\blacktriangledown$),
%    (\raisebox{\depth}{\rotatebox[origin=c]{180}{$\triangle$}}),
    and where F1 is highest ($\blacksquare$).
\textbf{Right:} F1 for each method, as well as the confidence threshold
for the marginal PR curve's max-F1 point. 
For path length 
\textbf{\textcolor[HTML]{1B9E77}{1}}, Greedy and MC-MAP F1 are the same as labeled attachment score (LAS).
 }
\end{figure*}}

\subsection{Monte Carlo Syntax Marginal (MCSM) inference for structure queries} \label{s:squery}

Beyond entire tree structures,
parse posteriors also define
marginal probabilities of particular events in them.
Let $f(y) \rightarrow \{0,1\}$ be a boolean-valued \emph{structure query}
function of a parse tree---for example, whether the tree contains a particular edge:
\[f(y) = 1\{\text{dobj(kill,Smith)} \in y\}\]
or more complicated structures, such as a length-2 dependency path:
%\begin{align*}
%&f(y) = \\
% &1\{\text{nsubj(kill},\text{cop}) \land \text{dobj}(\text{kill},\text{Smith}) \in y\}.
% \end{align*}
\[f(y) = 1\{\text{nsubj(kill},\text{cop}) \land \text{dobj}(\text{kill},\text{Smith}) \in y\}.\]
More precisely, these queries are typically formulated to check for edges between specific tokens,
and may check tokens' string forms.

Although $f(y)$ is a deterministic function, 
since the parsing model is uncertain of the correct parse,
we find
the \emph{marginal probability}, or expectation,
of a structure query by integrating out the posterior parse distribution---that is, 
the predicted probability that the parse has 
the property in question:
\begin{align}
p(f(y) \mid x) &= \sum_{y \in \mathcal{Y}(x)} f(y)\ p(y \mid x) 
\label{e:marginalproblem}
\\
\approx \tilde{p}(f(y) \mid x) &
%= \frac{1}{S} \sum_{s=1}^S f(y^{(s)})
 = \sum_{y \in \tilde{\mathcal{Y}}(x)} f(y) \frac{c(y)}{S}. \label{e:mcest}
\end{align}
Eq.\ \ref{e:marginalproblem} is the expectation with regard to the model's true probability distribution ($p$)
over parses from the domain of all possible parse trees $\mathcal{Y}(x)$ for a sentence,
while Eq.\ \ref{e:mcest} 
 is a Monte Carlo estimate of the query's marginal probability---the fraction of parse tree samples where the structure query is true.
% , here written in terms
% the Monte Carlo empirical distribution over parses ($\tilde{p}$),
% where its domain is the unique set of parse samples $\tilde{\mathcal{Y}}(x)$.
% \bocomment{should we call this the ``support'' of the MC dist, as in the set of parses with non-zero prob? a kinda technical distinction.}
% ---and $c(y) \equiv \sum_{s}^S 1\{y=y^{(s)}\}$ is the count of a parse tree within the sample.
We use this simple method for all inference in this work,
though importance sampling
\cite{Dyer2016RNNG}, particle filters \cite{Buys2015Deps},
or diverse k-best lists \cite{Zhang2014Diversity}
could support more efficient inference in future work.

\subsection{Probabilistic inference for dependencies: related work}
\label{relatedwork}
Our transition sampling method aims to be an easy-to-implement algorithm
for a highly performant class of dependency models,
that conducts exact probabilistic inference for arbitrary structure queries
in a reasonable amount of time.  A wide range of alternative methods
have been proposed for dependency inference
that cover some, but perhaps not all, of these goals.

For transition-based parsing,
beam search is a commonly used inference method that tries to look beyond a single structure.
Beam search can be used to yield an approximate $K$-best list
by taking resulting structures on the beam, though there are no theoretical guarantees about the result,
and runtime is no better than the transition sampler.\footnote{Loosely,
if it takes
 $N$ transitions to complete a parse, and $B$ possible actions at each transition
must be evaluated, our method evaluates $KNB$ actions
to obtain $K$ trees.
Beam search evaluates a similar number of actions when using a $K$-sized beam,
but also requires non-parallelizable management of the beam's priority queue.}
\citet{Finkel2006Pipeline}
further discuss tradeoffs between beam search and sampling, and find they 
give similar performance when propagating named entity recognition 
and PCFG parse information
to downstream tasks.

Graph-based parsers are the major alternative modeling paradigm
for dependency parsing;
instead of a sequence of locally normalized decisions,
they directly parameterize an entire tree's globally normalized probability.
Parse samples could be drawn from a graph-based model
via
Markov chain Monte Carlo
\cite{Zhang2014GibbsDeps}, which is asymptotically correct,
but may require a large amount of time to obtain non-autocorrelated parses.
A range of methods address inference
for specific queries in graph-based models---for example,
edge marginals for edge-factored models via the matrix-tree theorem
\cite{Koo2007MatrixTree},
or approximate marginals with loopy belief propagation
\cite{Smith2008Deps}.\footnote{These papers 
infer marginals to support parameter learning,
but we are not aware of previous work that directly analyzes
or uses dependency parse marginals.}
By contrast, our method is guaranteed to give
correct marginal inferences for arbitrary, potentially long-distance, queries.

Given the strong performance of graph-based parsers in the single-structure prediction setting
(e.g.\ \citet{zeman-EtAl:2017:K17-3,dozat2017stanford}),
it may be worthwhile to further explore probabilistic inference for these models.
For example, \citet{niculae2018sparsemap} present an inference algorithm
for a graph-based parsing model
that infers a weighted, sparse set of highly-probable parse trees,
and they illustrate that it can infer syntactic ambiguities similar to Figure~\ref{tels}.

Dynamic programming for dependency parsing, as far as we are aware,
has only been pursued for single-structure prediction (e.g.\ \citet{huang-sagae:2010:ACL}),
but in principle could be generalized to calculate local structure query marginals
via an inside-outside algorithm,
or to sample entire structures through an inside-outside sampler
\citep{eisner:2016:SPNLP}, which \citet{Finkel2006Pipeline} use to propagate parse uncertainty 
for
downstream analysis.

\section{Exploratory error analysis via whole-tree entropy calculations
} \label{s:eda}

 \begin{table*}[t]
\small
\centering
\begin{tabular}{p{10cm}ccc}
Sentence & Domain Size & Top 3 Freq. & Entropy \\ 
\toprule
\toprule
In Ramadi , there was a big demonstration . & 3 & [ 98, 1, 1 ] & 0.112 \\ \hline
US troops there clashed with guerrillas in a fight that left one Iraqi dead . & 40 & [ 33, 11, 6 ] & 2.865 \\ \hline
The sheikh in wheel - chair has been attacked with a F - 16 - launched bomb . & 98 & [ 2, 2, 1 ] & 4.577 \\ \hline
\end{tabular}
\caption{Example sentences from the UD development set and summaries of their Monte Carlo parse distributions.
\emph{Domain Size} gives $|\tilde{\mathcal{Y}}_{100}|$, the number of unique parse structures in 100 samples. \emph{Top 3 Freq.} gives the frequencies of the 3 most probable structures in $\tilde{\mathcal{Y}}_{100}$. \emph{Entropy} is calculated according to Eq. \ref{ent-eq};
its upper bound is $\log(100)=4.605$.}
\label{ent-ex}
\end{table*}
\normalsize

%\kkcomment{I feel like these two paragraphs are explained in the introduction. They feel a little redundant here.}
%We first explore uncertainties in model predictions.  
%Since ambiguity is endemic in natural language syntax, we expect a probabilistic parser 
%will often be unable to resolve ambiguities, and ideally, such uncertainty should be 
%represented in its marginal inferences.  
In this section we directly explore the model's intrinsic uncertainty,
while \S\ref{s:calib} conducts a quantitative analysis of model uncertainty compared to gold standard structures.
%Many sentences exhibit structural ambiguities, even short ones such as ``She saw the man with the telescope" (Fig.~\ref{tels}), where the attachment of the prepositional phrase is ambiguous (as seen in the two \emph{nmod} edges). Depending on the context, one reading may be preferred over the other; however, a parser is generally forced to immediately resolve the ambiguity by selecting one parse structure, discarding any others, and is unable to pass on the uncertainty downstream. 
 %\scomment{some notes here about passing on parsing uncertainty to downstream parsing instead of requiring immediate choice.}
%\scomment{how to tie this together? something like, beyond structural ambiguities, analyzing parse uncertainty can yield useful insights that typical error analysis approaches cannot do.}
%For a given set of $S$ parse samples yielding a collection of parse structures $\mathcal{Y}$, we can investigate parse uncertainty by calculating the entropy of the parse distribution:
%\begin{align}
%h(p) &= - \sum\limits_{y \in \mathcal{Y}} \tilde{p}(y) \log \tilde{p}(y) \nonumber \\
%&= - \sum\limits_{y \in \mathcal{Y}} \frac{c(y)}{S} \log \frac{c(y)}{S} \label{ent-eq}
%\end{align} 
Parse samples are able to both pass on parse uncertainty and yield useful insights that typical error analysis approaches cannot.
For a sentence $x$, we can calculate the \emph{whole-tree entropy}, the model's uncertainty of whole-tree parse frequencies 
in the samples:
\begin{align}
H(p) &= - \sum\limits_{y \in \mathcal{Y}(x)} p(y\mid x) \log p(y\mid x) \nonumber \\
\approx H(\tilde{p}) & 
= - \sum\limits_{y \in \tilde{\mathcal{Y}}(x)} 
\frac{c(y)}{S} \log \frac{c(y)}{S}. \label{ent-eq}
\end{align}

\noindent Since this entropy estimate is only based on an $S$-sample approximation of $p$,
it is upper bounded at $\log(S)$ in the case of a uniform MC distribution.
Another intuitive measure of uncertainty is simply the number of unique parses, that is, the cardinality of the MC distribution's domain $(|\tilde{\mathcal{Y}}|)$;
this quantity is not informative for the true distribution $p$, but in the MC distribution
it is intuitively upper bounded by $S$.\footnote{Shannon entropy, domain support cardinality,
and top probability ($\max_{y \in \tilde{Y}} \tilde{p}(y)$),
which we show in Table~\ref{ent-ex},
are all instances of the more general Renyi entropy \cite{Smith2007Renyi}.}
%\kkcomment{Does the ``MC" notation flow with the previous sections?}

\begin{figure}
\centering
\includegraphics[width=\linewidth]{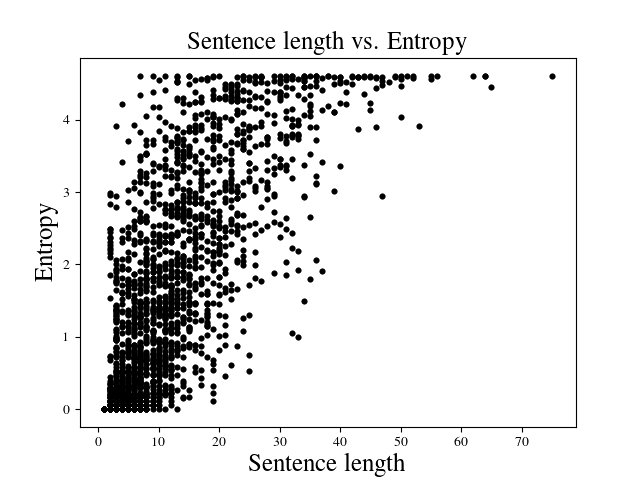}
\caption{Length of sentences (in number of tokens in UD development set) against entropy (100 samples per sentence).}
\label{len-ent}
\end{figure}

We run our dependency sampler on the 2002
sentences in the Universal Dependencies 1.3 English
Treebank development set, generating 100 samples per sentence; Table \ref{ent-ex} shows example sentences along with $|\tilde{\mathcal{Y}}|$ and entropy statistics for each sentence. We find that in general, as sentence length increases, so does the entropy of the parse distribution (Fig.~\ref{len-ent}). Moreover, we find that entropy is a useful diagnostic tool. For example, 7\% of sentences in the UD development corpus with fewer than 15 tokens and $H(\tilde{p}) \ge 2$ exhibit uncertainty around the role of `-' (compare \emph{Sciences - principally biology} and \emph{thought-provoking}), and another 7\% of such sentences exhibit uncertainty around `s' (potentially representing a plural or a possessive). 

\prfigure

\section{Monte Carlo Syntax Marginals for partial dependency parsing} \label{s:pr}

Here we examine the utility of marginal inference for predicting
parts of dependency parses,
using the UD 1.3 Treebank's English development set to evaluate.\footnote{UD 1.3 is the
UD version 
that this parsing model is most similar to:
\url{https://mailman.stanford.edu/pipermail/parser-user/2017-November/003460.html}}

\subsection{Greedy decoding} \label{ss:dataset}

Using its off-the-shelf pretrained model with greedy decoding,
the CoreNLP parser 
achieves 80.8\% labeled attachment score (LAS).
LAS is equivalent to both the precision and recall of predicting 
(rel,gov,child) triples
in the parse tree.\footnote{LAS is typically defined as proportion of tokens whose governor (and relation on that governor-child edge) are correctly predicted; this is equivalent to precision and recall of edges if all observed tokens are evaluated.
If, say, punctuation is excluded from evaluation, this equivalence does not hold; in this work we always use all tokens for simplicity.}
 
\subsection{Minimum Bayes risk (MBR) decoding}
A simple way to use marginal probabilities for parse prediction
is to select, for each token, the governor and relation that has the highest marginal probability.
This method gives a minimum Bayes risk (MBR) prediction of the parse,
minimizing the model's expected LAS with regards to local uncertainty;
similar MBR methods have been shown to improve accuracy in
tagging and constituent parsing (e.g.\ \citet{Goodman1996MBR,Petrov2007Parsing}).
This method yields 81.4\% LAS, outperforming greedy parsing, though
it may yield a graph that is not a tree.
 
%analyze.py
%MBR ACC 0.814259583267
%BRIER 0.173505686337

 \subsection{Syntax marginal inference for dependency paths} \label{ss:syntax-marg}
 \noindent
An alternative view on dependency parsing
is to consider what structures are needed for downstream applications.
One commonly used parse substructure is the \emph{dependency path} between two words,
which is widely used in unsupervised lexical semantics \citep{Lin2001DIRT},
distantly supervised lexical semantics \citep{Snow2005Hypernym},
relation learning \citep{Riedel2013Universal}, and
supervised semantic role labeling \citep{Hacioglu2004SRL,Das2014Semafor},
as well as applications in economics \citep{Ghose2007Reviews},
political science \citep{OConnor2013IR},
biology \citep{Fundel2006RelEx},
and the humanities \citep{Bamman2013Personas,Bamman2014Literary}.

In this work, we consider a dependency path to be a set of edges from the dependency parse;
for example, a length-2 path $p = \{\text{nsubj}(3,1), \text{dobj}(3,4)\}$ connects tokens 1 and 4.
Let $\mathcal{P}_d(y)$ be the set of all length-$d$ paths from a parse tree $y$.\footnote{Path construction may traverse both up and down directed edges; we represent a path as an edge set to evaluate its existence in a parse. A path may not include the same vertex twice. The set of all paths for a parse includes all paths from all pairs of vertexes (observed tokens and ROOT). %, where such a path exists.
}
Figure~\ref{f:pr}'s ``Greedy'' table column displays the F-scores for the precision and recall
of retrieving $\mathcal{P}_d(y^{(gold)})$ from the prediction $\mathcal{P}_d(y^{(greedy)})$
for a series of different path lengths.  $\mathcal{P}_1$ gives individual edges, and thus
is the same as LAS (80.8\%).  Longer length paths see a rapid decrease in performance; even length-2 paths
are retrieved with only $\approx 66\%$ precision and recall.\footnote{For length 1 paths, precision and recall are identical; this does not hold for longer paths, though precision and recall from a single parse prediction are similar.}
We are not aware of prior work that evaluates dependency parsing beyond single edge or whole sentence accuracy.

We define \emph{dependency path prediction} as the task of predicting a set of dependency paths for a sentence; the paths do not necessarily have to come from the same tree,
nor even be consistent with a single syntactic analysis.
We approach this task with our Monte Carlo syntax marginal method, 
by predicting paths from the transition sampling parser.
Here we treat each possible path as a structure query ($\S\ref{s:squery}$)
and return all paths whose marginal probabilities are at least threshold $t$.
Varying $t$ trades off precision and recall.

We apply this method to 100 samples per sentence in the UD treebank.
When we take all length-1 paths that appear in every single sample (i.e., estimated marginal probability $1.0$),
precision greatly increases to $0.969$, while recall drops to $0.317$ (the top-left point 
on Figure~\ref{f:pr}'s 
\textbf{\textcolor[HTML]{1B9E77}{teal}}
length-1 curve.)
We can also accommodate applications which may prefer to have a higher recall:
predicting all paths with at least $0.01$ probability
results in $0.936$ recall (the bottom-right point on the curve in Figure~\ref{f:pr}).\footnote{
% individual prec/rec numbers from
%> d=readfile("edgepred.tlen=1.100sample")
%> pp=performance(prediction(d$V2,d$V3),'prec','rec')
The 6.4\% of gold-standard edges with predicted 0 probability often
correspond to inconsistencies in the formalism standards between the model and UD;
for example, 0.7\% of the gold edges are `name' relations among words in a name, 
which the model instead analyzes as `compound'.
Inspecting gold edges' marginal probabilities helps error analysis, 
since when one views a single predicted parse,
it is not always clear whether observed errors are systematic, or a fluke for that one instance.}
% grep 'EDGEPRED 0 1 ' edgepred.tlen=1.100sample | less
%~/Desktop/parsemar/parsemar/ud_parses % grep AAA edgepred.tlen=1.100sample|awk '$3{ngold+=1} $2==0 && $3{missed+=1} END{print "gold total:",ngold, "  missed:", missed, missed/ngold}'      
%gold total: 25148   missed: 1609 0.0639812
%~/Desktop/parsemar/parsemar/ud_parses % grep AAA edgepred.tlen=1.100sample|grep 'name('|awk '$3{ngold+=1} $2==0 && $3{missed+=1} END{print missed/25148}'
%0.00731669

This marginal path prediction method dominates the greedy parser: for length-1 paths, 
there are points on the marginal decoder's PR curve that achieve both higher precision and recall
than the greedy decoder, giving F1 of 82.4\% when accepting all edges with marginal probability at least $0.45$.
Furthermore, these advantages are more prominent for longer dependency paths.
For example, for length-3 paths, the greedy parser only achieves 50.6\% F1, while the marginal parser
improves a bit to 55.0\% F1; strikingly, it is possible to select high-confidence paths to get much higher 90.1\% precision (at recall 11.6\%, with confidence threshold $t=0.95$).
%> d=readfile("edgepred.tlen=3.100sample")
%> pp=performance(prediction(d$V2,d$V3),'prec','rec')
Figure~\ref{f:pr} also shows the precision/recall points on each curve for thresholds $t=0.9$ and $t=0.1$.

We also evaluated the MC-MAP single-parse prediction method (\S\ref{s:mcmap}),
which slightly, but consistently,
underperforms the greedy decoder at all dependency lengths.
More work is required to understand whether this is is an inference or modeling problem:
for example, we may not have enough samples to reliably predict a high-probability parse;
or, as some previous work finds in the context of beam search,
the label bias phenomenon
in
this type of locally-normalized transition-based parser 
may cause it to assign higher probability to non-greedy analyses 
that in fact have lower linguistic quality
\cite{Zhang2012Deps,Andor2016Parseface}.

\section{Calibration}\label{s:calib}

\begin{figure}[t]
\includegraphics[width=\linewidth,trim=0 0 0 0.7in]{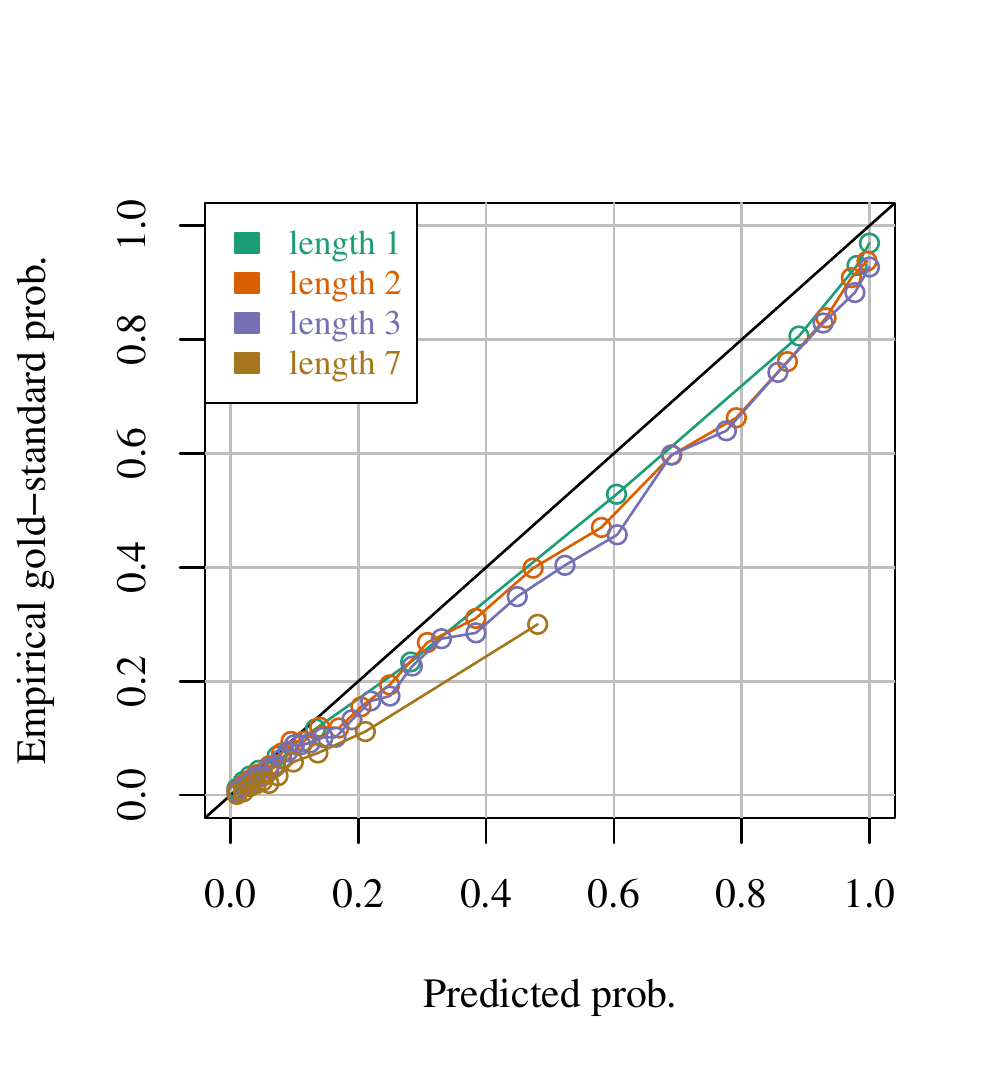}
\vspace{-0.4in}
\caption{\label{f:calib} Calibration curves for marginal predictions for
several path lengths.  Predictions below the $y=x$ line indicate overconfidence.
Points denote the average predicted probability versus empirical (gold) probability
among predicted paths within each dynamically allocated bin.
}
\end{figure}

The precision-recall analysis shows that the predicted marginal probabilities
are meaningful in a ranking sense, but we can also ask whether they are meaningful in a sense
of \emph{calibration}:
predictions are calibrated if, among all structures with predicted probability $q \pm \epsilon$,
they exist in the gold parses with probability $q$.  That is, predictions with confidence $q$ have precision $q$.\footnote{This is a \emph{local} precision,
as opposed to the more usual tail probability of
measuring precision of all predictions \emph{higher} than some $t$---the integral of local precision.
For example, Figure~\ref{f:pr}'s length-1 $t=0.9$ 
precision of $0.942$ (\textbf{\textcolor[HTML]{1B9E77}{$\triangle$}})
is the average $y$ value of several rightmost bins in Figure~\ref{f:calib}.
This contrast corresponds to \citet{Efron2010EB}'s dichotomy of local versus global false discovery rates.}
If probabilities are calibrated,  that implies expectations with regard to their distribution are unbiased,
and may also justify intuitive interpretations of probabilities in exploratory analysis (\S\ref{s:eda}).
Calibration may also have implications for joint inference, EM, and active learning methods that
use confidence scores and confidence-based expectations.

We apply \citet{Nguyen2015Calib}'s adaptive binning method to analyze the calibration of 
structure queries from an NLP system, by taking the domain of all seen length-$d$ paths 
from the 100 samples' parse distribution for the treebank,
grouping by ranges of predicted probabilities to have at least 5000 paths per bin,
to ensure stability of the local precision estimate.\footnote{This does not include gold-standard paths with zero predicted probability.
%The algorithm constructs 
%boundaries from left to right, so the rightmost bucket can have less than 5000 items. 
As \citeauthor{Nguyen2015Calib} found for sequence tagging and coreference, we find the prediction distribution is heavily skewed to near 0 and 1, necessitating adaptive bins,
instead of fixed-width bins, for calibration analysis \citep{Niculescu2005Calibration,Bennett2000NB}.
}

We find that probabilities are reasonably well calibrated, if slightly overconfident---Figure~\ref{f:calib}
shows the average predicted probability per bin, compared to how often these paths appear in the gold standard (local precision).
For example, for edges (length-1 paths),
predictions near 60\% confidence (the average among predictions in range $[0.42,0.78]$)
correspond to edges that are actually in the gold standard
tree only 52.8\% of the time.
The middle confidence range has worse calibration error, and longer paths perform worse.
Still, this level of calibration seems remarkably good,
considering
there was no attempt to re-calibrate predictions \citep{Kuleshov2015Calib}
or to use a model that specifically parameterizes
the energy of dependency paths \citep{Smith2008Deps,Martins2010Turbo}---these predictions are simply 
a side effect of the overall joint model
for incremental dependency parsing.

%In the remaining sections, we investigate the use of parse samples for rule- (\S\ref{s:rules}) and machine learning-based (\S\ref{s:ml}) information extraction systems.

%Longer dependency paths display increasing overconfidence.
%We also show a more extreme case of length-7 dependency paths;
%only 394 of the 2002 sentences have at least one such path.
%%~/Desktop/parsemar/parsemar/ud_parses % grep 'pred paths support' edgepred.tlen=7.100sample | awk '$1>0'|wc -l     
%%     394
%It is very difficult for the model to give high predicted probabilities to such paths;
%among all samples in all sentences, there are 4734 length-7 paths
%in the right-most bucket with predicted probabilities from 0.29 to 1.0
%(and only 6 with probability 1.0).
%Their average predicted probability is 48\%, while they are empirically in the gold standard only 30\% of the time.
%

%\begin{table}[t]
%  \centering
%      \begin{tabular}{l r r}
%               & Train  & Test    \\
%      \toprule
%      \toprule
%      Positive sentences & 11,274 & 6,132   \\
%      Negative sentences & 121,559 & 62,793 \\
%      Total sentences & 132,833 & 68,925
%  \end{tabular}
%  \caption{Data statistics for Police Fatality experiments  \label{t:pf-stats}}
%\end{table}

%%%%%%%%%%%%%%%%%%%%%%%%%%%%%%%%
\newcommand{\pffig}{\begin{figure}[t]
\begin{minipage}{2.1in}
\includegraphics[width=\linewidth]{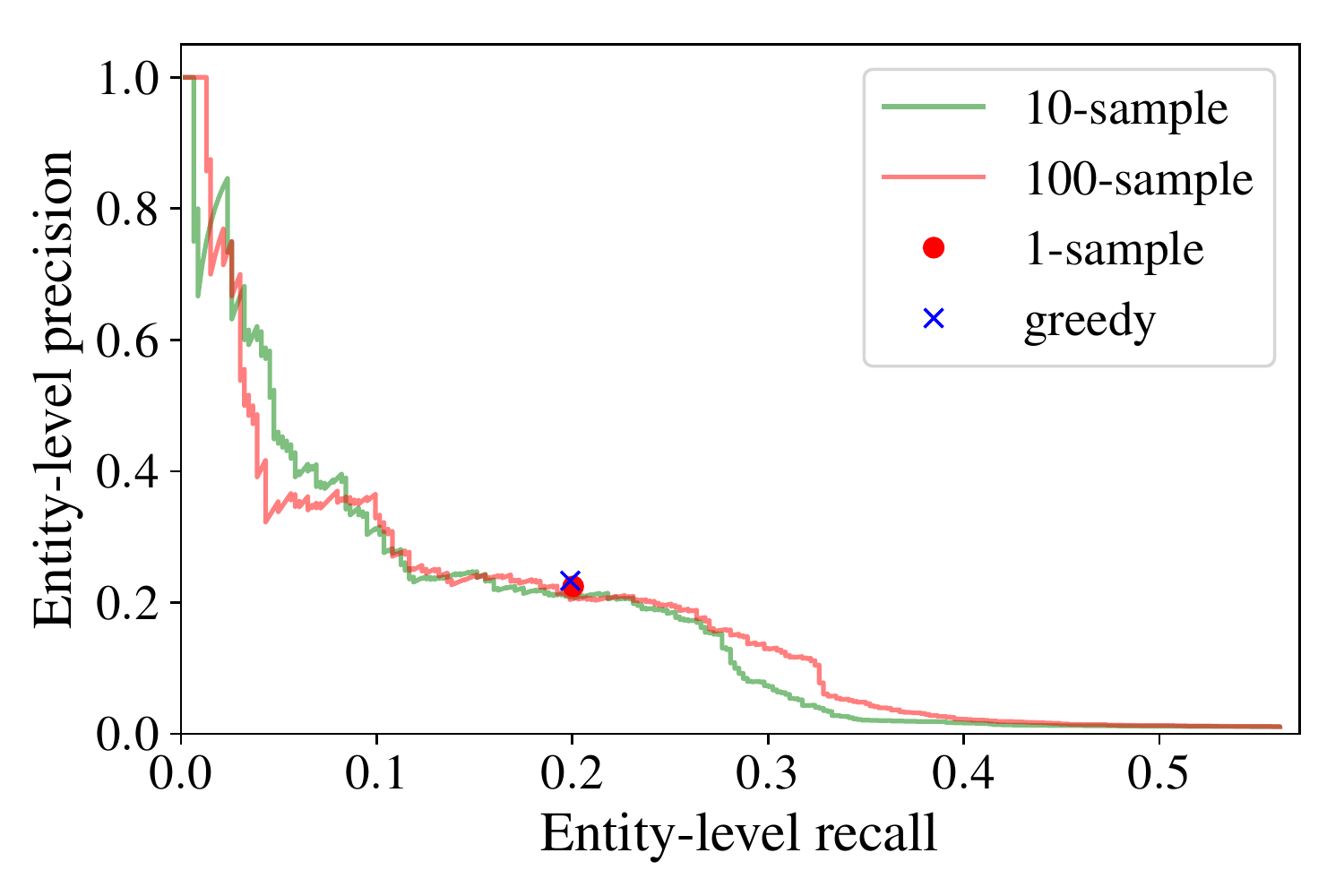}
\end{minipage}
\hspace{-0.1in}
\begin{minipage}{0.5in}
{\footnotesize
      \begin{tabular}{@{}ll@{}}
              Method & F1    \\
      \cmidrule(r){1-1} \cmidrule(l){2-2}
      RPI-JIE & 0.170\\
      Greedy & 0.215\\
      1 samp. & 0.212\\
      10 samp. & 0.219 \\
      100 samp. & \textbf{0.222}\\ 
  \end{tabular}
  }
\end{minipage}
\caption{\textbf{Left:} Rule-based entity precision and recall for police fatality victims,
with greedy parsing and Monte Carlo inference. \textbf{Right:} F1 scores for RPI-JIE, Greedy, and 1-sample methods, and maximum F1 on PR curve for probabilistic (multiple sample) inference.
\label{f:pf-prrec}
}
\end{figure}}

\section{Probabilistic rule-based IE: classifying police fatalities} \label{s:rules}
Supervised learning typically gives the most accurate information extraction or semantic parsing systems,
but for many applications where training data is scarce, 
\citet{Chiticariu2013LongLiveRule} argue that rule-based systems are
useful and widespread in practice,
despite their neglect in contemporary NLP research.
Syntactic dependencies are a useful abstraction 
with which to write rule-based extractors,
but they can be brittle due to errors in the parser.
We propose to integrate over parse samples to infer a \emph{marginal} probability of a rule match,
increasing robustness and allowing for precision-recall tradeoffs.

\subsection{Police killings victim extraction}
We examine the task of extracting the list of names of persons killed by police
from a test set of web news articles in Sept--Dec 2016.
We use the dataset released by \citet{keith2017identifying},
consisting of 24,550 named entities $e \in \mathcal{E}$ and sentences from noisy web news text extractions
(that can be difficult to parse), each of which contains at least one $e$ (on average, 2.8 sentences/name)
 as well as keywords for both police and killing/shooting. 
The task is to classify whether a given name is a person who was killed by police,
given 258 gold-standard names that have been verified by journalists.  

%% BTO: i added but too verbose.
%In a sentence prediction framework, a mention-level classifier determines whether a particular sentence $i$
%asserts the mentioned person $e$ was killed by police; a hard classifier $f(x_i,e)$ returns 1 or 0, and an aggregate entity-level prediction can be determined true if at least one sentence asserts the event,
%\begin{align} \text{HardClassif}(e) &= \bigvee_{i \in \mathcal{M}(e)} f(x_i,e),  \label{e:hardor} \end{align}
%or, if we have access to a sentence-level probabilistic inference $q(x_i,e) \in [0,1]$,
%a probabilistic entity-level prediction can be inferred as the probability at least one of the sentences
%asserts the event---that is, the NoisyOR rule \cite{TODO cites from before},
%\begin{align} \text{SoftClassif}(e) &= 1-\prod_{e \in \mathcal{M}(e)} (1-q(x_i,e)). \label{e:noisyor} \end{align}

\subsection{Dependency rule exractor} \label{ss:rule-based}
\citeauthor{keith2017identifying} present a baseline rule-based method
that uses \citet{li2014incremental}'s off-the-shelf RPI-JIE ACE event parser
to extract (event type, agent, patient) tuples from sentences,
and assigns $f_{\text{JIE}}(x_i,e)=1$ iff the event type was a killing,
the agent's span included a police keyword,
and the patient was the candidate entity $e$.
An entity is classified as a victim if at least one sentence is classified as true,
%$f_{\text{JIE}}=1$
%Entity classification via the OR rule (\ref{e:hardor}) 
resulting in a 0.17 F1 score (as reported in previous work).\footnote{This measures
recall of the entire gold-standard victim database, though the corpus only includes 57\% of the victims.}

We define a similar syntactic dependency rule system using a dependency parse as input:
our extractor $f(x,e,y)$ returns 1 iff
the sentence has a killing keyword $k$,\footnote{Police and killing/shooting keywords are from \citeauthor{keith2017identifying}'s publicly released software.}
 which both
 \begin{enumerate}
 \itemsep0em
 \item has an agent token $a$ (defined as, governed by \emph{nsubj} or \emph{nmod})
which is a police keyword, or $a$ has a (\emph{amod} or \emph{compound}) modifier that is a police keyword;
and,
\item has a patient token $p$ (defined as, governed by \emph{nsubjpass} or \emph{dobj})
contained in the candidate name $e$'s span.
\end{enumerate}
Applying this $f(x,e,y)$ classifier to greedy parser output, it performs better than the RPI-JIE-based
rules (Figure~\ref{f:pf-prrec}, right),
perhaps because it is better customized for the particular task.

\pffig

Treating $f$ as a structure query, we then use our Monte Carlo marginal inference (\S\ref{s:model})
method to calculate the probability of a rule match for each sentence---that is,
the fraction of parse samples where $f(x,e,y^{(s)})$ is true---and infer the entity's probability
with the \emph{noisy-or} formula \citep{Craven1999Bio,keith2017identifying}.
This gives soft classifications for entities.

\subsection{Results}
The Monte Carlo method achieves slightly higher
F1 scores once there are at least 10 samples (Fig.~\ref{f:pf-prrec}, right).
More interestingly, 
the soft entity-level classifications also allow for precision-recall tradeoffs (Fig.~\ref{f:pf-prrec}, left),
which could be used to prioritize the time of human reviewers updating the victim database (filter to higher precision), or help ensure victims are not missed (with higher recall).
We found the sampling method retrieved several true-positive entities
where only a single sentence had a non-zero rule prediction at probability 0.01---that is, 
the rule was only matched in one of 100 sampled parses.
Since current practitioners are already manually reviewing millions of news articles
to create police fatality victim databases, 
the ability to filter to high recall---even with low precision---may be useful
to help ensure victims are not missed.

\subsection{Supervised learning}
Sampling also slightly improves supervised learning for this problem.
We modify \citeauthor{keith2017identifying}'s logistic regression model based on
a dependency path feature vector $f(x_i,y)$,
instead creating feature vectors that average over
multiple parse samples ($E_{\tilde{p}(y)}[f(x_i,y)]$) at both train and test time.
With the greedy parser, the model results in 0.229 F1;
using 100 samples slightly improves performance to 0.234 F1.
%\bocomment{I rounded-nearest 0.2338 to 0.234.  katie, if you were using round-down to go down to 3 digits elsewhere, please change to be consistent.}

%to 0.2356 and 0.2338, respectively.
%\bocomment{moved here. may cut.}
%Using the police fatality data described in \S\ref{s:rules}, we extracted pre-averaged features from the 10- and 100-sample parse sets for each sentence used them in logistic regression. We compared to the baseline entity-level area under the precision recall curve (AUPRC) and F1 scores of 0.117 and  respectively \cite{keith2017identifying}. Introducing parse samples slightly increased performance to 0.1177 AUPRC and 
%0.2356 F1 for 10 samples and 0.1183 AUPRC and 0.2338 F1 for 100 samples. 
%%\kkcomment{Is there any positive spin here? About how how we could have even better results on an easier task?  Or how noisy the data and entity-level aggregation is for this task? Or should we just put this as a footnote?}

\section{Semantic role assignment} \label{s:ml}

%In addition to using (MCSM) inference in rule-based systems, we explore two approaches for using parse samples as features in machine learning pipelines.
%
%\textbf{Approach 2: Joint model.} 
%If we define the IE model based only on a single parse output,
%then the full model requires integrating out all possible parses, which we again approximate with the Monte Carlo expectation:
%\begin{align}
%%p(z=1 \mid y) &= \sigma( \theta\transpose f(y) ) \\
%p(z =1 \mid x) &= \sum_{y \in \mathcal{Y}(x)} p(z =1 \mid y) p(y \mid x) \\
%& \approx \frac{1}{S} \sum_{s=1}^S p(z = 1\mid y^{(s)}) 
%\end{align}
%Note there is no $g(x)$ in this approach. Instead, the parse structures are treated as a latent variable, and hence the modeler is no longer required to decide ahead of time how to aggregate information across the parse posterior. This is particularly desirable for more complex structural computations, such as recurrent neural networks on a dependency parse.
%%and we don't have to decide ahead of time how to aggregate information across the parse posterior.
%%This could be nice if we wanted more complicated structural computations, like recurrent neural networks on a dependency parse 
%\scomment{Brendan: can you do these citations? (CITE -- JHU stuff by ... violet? eisner? dredze? others?)}
%%which makes sense only in the context of individual parses.
%

\label{ss:srl}
Semantic role labeling (SRL), the task to predict argument structures \cite{gildea2002automatic},
is tightly tied to syntax, and previous work has found it beneficial to 
conduct it with joint inference with constituency parsing, such as
with top-$k$ parse trees \cite{Toutanova2008SRL} or parse tree samples \cite{Finkel2006Pipeline}.
Since \S\ref{s:pr} shows that Monte Carlo marginalization improves dependency edge prediction,
we hypothesize dependency sampling could improve SRL as well.

SRL includes both identifying argument spans, and assigning spans to specific semantic role labels (argument types).
We focus on just the second task of semantic role assignment:
assuming argument spans are given, to predict the labels.
We experiment with English OntoNotes v5.0 annotations \cite{Weischedel2013OntoNotes}
according to the CoNLL 2012 test split \cite{pradhan2013towards}.
%the version of the data in the LDC OntoNotes release.}
We focus only on predicting among the five core arguments (Arg0 through Arg4) and ignore spans with gold-standard adjunct or reference labels. % (e.g.\ AM-ADV, R-A1).
%The semantic model for label $z$ given the parse, $p(z \mid y)$,
%is simply the conditional probability 
We fit a separate model for each predicate\footnote{That is,
for each unique 
(lemma, framesetID) pair, such as (view, view-02).}
among the 2,160 predicates that occur at least once in both the
training and test sets (115,811 and 12,216 sentences respectively).
%from 115,811 training-time sentences and 12,216 test-time sentences.
%This results in in 115,811 sentences with 5,287 unique predicate tuples for training and
%12,216 sentences with 2,160 unique tuples for testing. 

Our semantic model of label $z_t \in \{A0..A4\}$
for argument head token $t$ and predicate token $p$, 
$p_{\text{sem}}(z_t \mid p,y)$,
is simply the conditional probability of the label, conditioned on
 $y$'s edge between $t$ and $p$ if one exists.\footnote{The dataset's argument spans must be reconciled with predicted parse structures to define the argument head $t$;
  90\% of spans are consistent with the greedy parser in that all the span's tokens have the same highest ancestor contained with the span, which we define as the argument head.
  For inconsistent cases,
  we select the largest subtree
  (that is, highest within-span ancestor common to the largest number of the span's tokens).
  It would be interesting 
  to modify the sampler to restrict to
  parses that are consistent with the span, as a form of rejection sampling.}
 (If they are not directly connected, the model instead conditions on a `no edge' feature.)
 Probabilities are maximum likelihood estimates from the training data's
 (predicate, argument label, path) counts,
 from either greedy parses, or averaged among parse samples.
To predict at test time,
the greedy parsing model simply uses $p(z_t \mid p, y^{(greedy)})$.
The Monte Carlo model, by contrast, treats it as a directed joint model
and marginalizes over syntactic analyses:
\begin{align}
p_{MC}(z_t \mid p,x) = \sum_{y \in \tilde{\mathcal{Y}}(x)} p_{\text{sem}}( z_t \mid p,y)\ \tilde{p}_{\text{syn}}(y \mid x). \nonumber
\end{align}
 %We discard sentences that (1) do not have a predicate identified in the OntoNotes annotation, (2) have a predicate at test time that was not seen at training time, or (3) have a feature for a predicate that was seen at test time but not training time.\footnote{We also experimented with calculating accuracy by backing off to the most predicted argument if we encountered these ``null" features as test time. This resulted in small changes in accuracy to 0.501 for greedy and 0.526 for 100-sample. The expected non-null rate was also high with 0.981 for greedy and  0.996 for 100-sample.}
%As a baseline, for each predicate we predict the argument that was seen the most at training time. 

\noindent
The baseline accuracy of predicting the predicate's most common training-time argument label
yields 0.393 accuracy, and the greedy parser performs at 0.496.
The Monte Carlo method (with 100 samples) improves accuracy to 0.529
(Table~\ref{t:srl-results}).
Dependency samples' usefulness in this limited case suggests 
they may help systems that use dependency parses more broadly for SRL
\cite{Hacioglu2004SRL,Das2014Semafor}.

\begin{table}[t]
  \centering
  \footnotesize
      \begin{tabular}{l c}
      Method & Accuracy \\
	\cmidrule(lr){1-1} \cmidrule(lr){2-2}
      Baseline (most common) & 0.393 \\
      Greedy &0.496 \\
      MCSM method, 100-samples & \textbf{0.529}
  \end{tabular}
  \caption{Semantic role assignment accuracy on English OntoNotes v5.0. The baseline is for each unique predicate, predict the argument that was seen the most at training time. 
%  Accuracy rates discard null features. (See appendix for accuracy rates with backing off to the baseline without nulls).
   \label{t:srl-results}}
\end{table} 

\section{Conclusion}
In this work, we introduce a straightforward algorithm for sampling from the full joint distribution of a transition-based dependency parser. We explore using these parse samples to discover both parsing error and structural ambiguities. Moreover, we find that our Monte Carlo syntax marginal method not only dominates the greedy method for dependency path prediction (especially for longer paths), but also allows for control of precision-recall tradeoffs. Propagating dependency uncertainty 
can potentially help
a wide variety of semantic analysis and information extraction tasks.

\section*{Acknowledgments}
The authors would like to thank Rajarshi Das, Daniel Cohen, Abe Handler, Graham Neubig,
Emma Strubell,
and the anonymous reviewers for their helpful comments.

%\bibliography{naaclhlt2018}
\bibliography{brenocon,kkeith}
\bibliographystyle{acl_natbib}

\end{document}